\documentclass{article}

\usepackage{arxiv}

\usepackage[utf8]{inputenc} % allow utf-8 input
\usepackage[T1]{fontenc}    % use 8-bit T1 fonts
\usepackage{hyperref}       % hyperlinks
\usepackage{url}            % simple URL typesetting
\usepackage{booktabs}       % professional-quality tables
\usepackage{amsfonts}       % blackboard math symbols
\usepackage{nicefrac}       % compact symbols for 1/2, etc.
\usepackage{microtype}      % microtypography
\usepackage{lipsum}
\usepackage{graphicx}
\usepackage{amsmath}
\usepackage{amssymb}
\usepackage{multirow}
\usepackage{float}
\usepackage{makecell}
\usepackage{subfigure}
\usepackage{hyperref}
\hypersetup{
	colorlinks=true,
	linkcolor=blue,
	citecolor=blue,
	urlcolor=blue
}

\title{PCT-Prompt: A Prompt-Guided Transformer Framework for Dense Prediction Tasks in Point Clouds}

\author{
  Dejun Zhang\\
  China University of Geosciences\\
  \texttt{zhangdejun@cug.edu.cn} \\
   \And
  Yanzi Bai\\
  Li Auto Inc \\
  \texttt{baiyanzi123@cug.edu.cn} \\
  \And
  Yiqi Wu\footnote{Corresponding author}\\
  China University of Geosciences\\
  \texttt{wuyq@cug.edu.cn} \\
}
\begin{document}
	
\maketitle
\begin{abstract}
Standard Transformers have proven effective in point cloud object classification, but their performance in dense prediction tasks within complex scenes is often hindered by weak prior assumptions. To address this challenge, we propose PCT-Prompt, a novel framework that enhances standard Transformers by introducing a prompt-guided feature branch to improve performance in dense prediction tasks. The standard Transformer branch leverages pre-trained models for global feature extraction from point cloud data, serving as the backbone for processing high-level features. Meanwhile, the prompt-guided feature branch consists of two key components: a fine-grained feature extraction block that captures multi-scale geometric features using geometry-sensitive abstraction layer, along with the PnP-3D layer to integrate local context with global regularization. The second component, the prompt-refined feature learning block generates prompt tokens, which are subsequently refined through cross-attention mechanisms. Additionally, we introduce a prompt drop mechanism that progressively removes prompt information across Transformer layers, balancing local details and global consistency. Experimental results on the ShapeNetPart, S3DIS, and DALES datasets demonstrate that PCT-Prompt significantly improves the adaptability of standard Transformers to dense prediction tasks, achieving strong performance in real-world scenarios.
\end{abstract}

% keywords can be removed

\section{Introduction}
The widespread use of point cloud data in autonomous driving and robotics has significantly driven the development of deep learning methods specifically designed for point clouds. Unlike traditional deep learning architectures~\cite{qi2017pointnet++,wang2019dynamic,li2018pointcnn,choe2022pointmixer}, Transformers have emerged as particularly effective, excelling in handling sets and offering unparalleled adaptability to point cloud data. Point cloud Transformers are particularly proficient at capturing both local features and complex global relationships inherent in point clouds, which enables them to deliver superior performance in key tasks such as classification~\cite{sun2022fusing}, part segmentation~\cite{park2023self}, semantic segmentation~\cite{zhang2020pointwise} and registration~\cite{hu2022nrtnet}. The transformative advantages offered by Transformers have fueled innovation in point cloud data processing, driving notable advancements in both autonomous driving and robotics.

Transformer-based methods for point cloud processing often incorporate specific inductive biases, such as neighbor embedding, vector self-attention, and offset-attention. These biases are designed to enhance the perception of geometric structures and improve the ability to capture complex local features. Transformers that leverage these inductive biases are commonly referred to as \textbf{variant Transformers}, with notable examples including Point Transformer~\cite{zhao2021point} and Point Cloud Transformer~\cite{guo2021pct}. While these variant Transformers offer substantial improvements in dense prediction tasks, they may sacrifice generality and limit the effective representation of multi-modal features. In contrast, \textbf{standard Transformers}~\cite{yu2022point,pang2022masked,qi2023contrast,dong2022autoencoders,liu2022masked} provide several advantages, including ease of deployment, a flexible architecture, rich pre-training strategies, and the capacity to handle multi-modal data more effectively.

However, standard Transformers have several limitations: 
(1) Lack of inductive bias: without inductive biases, they require large amounts of training data for good generalization. These models~\cite{liu2022masked,yu2022point} are typically pre-trained on small datasets like ShapeNetPart~\cite{chang2015shapenet}, resulting in poor generalization. 
(2) Difficulty with multi-scale context: single-scale Transformers struggle to capture local context at different scales, which limits their performance in tasks involving varying object sizes.
(3) Dependence on initial self-attention: standard Transformers rely on initial self-attention mechanisms, restricting their performance in dense prediction tasks. Fine-tuning methods~\cite{tang2024point,zha2023instance} show some improvement in classification and part segmentation but still fall short for real-world scene segmentation.

We present PCT-Prompt, a novel framework that combines a standard Transformer branch with a prompt-guided feature branch. This approach leverages the strengths of standard Transformers while improving their performance on dense prediction tasks. The standard Transformer branch includes patch embedding and multiple Transformer blocks, which can load various pre-trained weights, providing a foundation for processing point cloud data and extracting global features. In parallel, the prompt-guided feature branch introduces two key components to enhance performance on dense prediction tasks:
(1) The Fine-grained Feature Extraction (FFE) block, which employs a hierarchical architecture to extract multi-scale geometric features using geometry-sensitive abstraction (GSA) layer, along with a plug-and-play layer, PnP-3D, that integrates local context fusion and global bilinear regularization.
(2) The Prompt-refined Feature Learning (PFL) block, which includes a prompt generator that dynamically transforms multi-scale geometric features into fine-grained prompt tokens. These tokens are then refined by a prompt refiner, which iteratively integrates both global and local features through cross-attention and shared-weight MLPs. 
Additionally, a prompt drop mechanism progressively removes prompt information across Transformer layers, enabling the model to focus on local geometric details while preserving global semantic consistency.

To summarize, this work makes the following contributions:
\begin{itemize}
	\item We propose PCT-Prompt, a framework combining standard Transformers with a prompt-guided feature branch for improved dense prediction tasks.
	\item We propose a Fine-grained Feature Extraction (FFE) block that leverages a hierarchical combination of the GSA and PnP-3D layers to capture multi-scale geometric features.
	\item We present a Prompt-refined Feature Learning (PFL) block that generates and refines prompt tokens, incorporating a prompt drop mechanism for balanced feature refinement.
	\item Experimental evaluations on ShapeNetPart, S3DIS, and DALES datasets show that PCT-Prompt significantly enhances the standard Transformer’s adaptability to dense prediction tasks.
\end{itemize}

The remainder of this article is organized as follows:
Section~\ref{sec::RelatedWork} reviews relevant literature.
Section~\ref{sec::PCT-Prompt} details the proposed PCT-Prompt framework, explaining its architecture and how the standard Transformer branch is enhanced with the prompt-guided feature branch for improved performance on dense prediction tasks.
Section~\ref{sec::Experiments} describes the datasets, training setup, evaluation metrics, experimental comparisons with state-of-the-art methods, and ablation studies.
Finally, Section~\ref{sec::Conclusion} provides the conclusion of the article.

\section{Related Work}
\label{sec::RelatedWork}
In this section, we provide an overview of related works in three key areas: deep learning-based point cloud processing, the application of transformers to point cloud data, and the emerging field of prompt tuning. We highlight the latest advancements in each domain, focusing on their relevance to current research and their role in addressing complex 3D perception challenges.

\subsection{Deep Learning for Point Cloud}
Point clouds are essential in various domains, including autonomous driving, robotics, and computer vision, due to their ability to represent 3D environments. 
They are commonly used in tasks such as classification~\cite{wu2025multimodal}, segmentation~\cite{zhang2026sam}, 3D reconstruction~\cite{tan2022coarse}, scene flow estimation~\cite{zhang2024bridging,yan2025flowst}, and cross-view geo-localization~\cite{xu2026weather}, which are crucial for understanding and interacting with 3D scenes. 
These tasks support applications like object recognition, scene reconstruction, and environmental mapping. 
Consequently, the development of deep learning methods for efficient and accurate point cloud processing has become a significant research focus.

Deep learning methods for point cloud processing can be broadly classified into three main categories: projection-based, voxel-based, and point-based approaches. 
Projection-based methods, such as PointCLIP~\cite{zhang2022pointclip}, and PointCLIPV2~\cite{zhu2023pointclip}, project point clouds onto multiple 2D views and utilize image feature extraction techniques to construct 3D representations. 
While these methods are efficient, they may suffer from occlusion and geometric loss. 
Voxel-based methods, including VoxNet~\cite{maturana2015voxnet}, OctNet~\cite{riegler2017octnet}, and O-cnn~\cite{wang2017cnn}, convert point clouds into voxel grids and apply 3D convolutions for feature extraction. 
These methods offer an effective way to handle large-scale data but can be memory-intensive and may lose resolution due to voxelization. 
Point-based methods, such as PointNet~\cite{qi2017pointnet}, PointNet++~\cite{qi2017pointnet++}, KPConv\cite{thomas2019kpconv}, Point Transformer~\cite{zhao2021point}, and PCT~\cite{guo2021pct}, process raw point cloud data directly, preserving its geometric structure. PointNet~\cite{qi2017pointnet} uses point-wise MLPs with global pooling to extract features, while PointNet++~\cite{qi2017pointnet++} employs a hierarchical approach for fine-grained local feature capture. KPConv~\cite{thomas2019kpconv} uses a continuous space point kernel to perform feature extraction, mimicking the behavior of convolutions on 3D data.

While projection-based and voxel-based methods offer simplified point cloud processing, they often face challenges like occlusion, geometric distortion, or high memory demands. 
In contrast, point-based methods retain the original geometry of point clouds, allowing for more precise and intuitive feature extraction. Building on this foundation, this paper adopts a point-based approach to extend the standard Transformer, enhancing its ability to capture multi-scale and detailed geometric features for dense prediction tasks.

\subsection{Transformer for Point Cloud}
Research on point cloud transformers can be categorized into two main directions: 
1) \textbf{Variant Transformers} with inductive biases tailored to point clouds~\cite{guo2021pct,zhao2021point,zhang2022patchformer,yang2025swin3d}; 
2) \textbf{Standard Transformers} adhering to the foundational goal of minimizing inductive bias~\cite{qi2023contrast,pang2022masked,yu2022point,dong2022autoencoders,liu2022masked}.

The first focuses on \textbf{variant transformers}, which incorporate inductive biases tailored to point clouds. 
For example, PCT~\cite{guo2021pct} enhances the model's geometric awareness by introducing inductive biases such as neighborhood embedding and offset attention, alongside the standard attention mechanism. 
Point Transformer~\cite{zhao2021point} integrates vector self-attention layers and position encoding to refine how the Transformer handles point clouds. Swin3D~\cite{yang2025swin3d} and Patchformer~\cite{zhang2022patchformer} further optimize computational efficiency by limiting attention mechanisms to local regions.
Although these methods show strong performance in point cloud tasks, the careful design of inductive biases introduces significant time costs, which can somewhat deviate from the original aim of minimizing inductive bias in Transformer models.

The second direction focuses on \textbf{standard transformers}, which aim to minimize inductive bias while leveraging pre-training strategies to enhance feature extraction capabilities. 
Point-BERT~\cite{yu2022point} extends the BERT~\cite{devlin2019bert} self-supervised learning approach to point clouds, achieving state-of-the-art results in classification and part segmentation through masked point modeling. MaskPoint~\cite{liu2022masked} introduces a discriminative pre-training framework that converts fuzzy masked point reconstruction into a more discriminative task, capturing rich features. 
Point-MAE~\cite{pang2022masked} employs random masking of point cloud patches for self-supervised pre-training, reconstructing unmasked patches to extract advanced latent features. 
ACT~\cite{dong2022autoencoders} and ReCon~\cite{qi2023contrast} further demonstrate the effectiveness of cross-modal learning, leveraging image and NLP Transformers for point cloud pre-training.

Minimizing inductive bias in standard Transformers has attracted considerable attention due to their ability to generalize across tasks. Building on this foundation, our study seeks to extend the pre-trained standard Transformer framework, enhancing its capabilities to tackle dense prediction tasks in complex scenes.

\subsection{Prompt Tuning}
Prompt tuning~\cite{su2022transferability}, originally developed for adapting pre-trained models to downstream tasks in natural language processing (NLP)~\cite{zhang2019attention}, has since been successfully extended to image, video, and multimodal domains. In NLP, GPT-3 demonstrated strong performance across various tasks through prompt-based few-shot learning, accelerating the use of prompt tuning. In image-related tasks, CLIP~\cite{radford2021learning} leverages contrastive learning to map images and text into a shared embedding space, enabling zero-shot learning through prompts for image classification and retrieval. Similarly, ALIGN~\cite{jia2021scaling} integrates vision and language information, optimizing multimodal tasks through prompts, thus improving image-text retrieval. VQGAN-CLIP~\cite{crowson2022vqgan} combines the Generative Adversarial Network (VQGAN)~\cite{esser2021taming} with CLIP~\cite{radford2021learning}, utilizing a pre-trained image-text joint encoder and natural language prompts for high-quality image generation and semantic editing.

In point cloud Transformer research, prompt tuning has also gained traction. For instance, Point-PEFT~\cite{tang2024point} combines point-granular prompts with a geometry-aware adapter, improving classification performance while maintaining a small parameter size via fine-tuning the Transformer pre-trained model. Point-MAE~\cite{pang2022masked} employs a masked autoencoder (MAE) method, using prompts to help the model reconstruct missing point cloud information, demonstrating strong results in point cloud reconstruction. IDPT~\cite{zha2023instance} introduces dynamic prompt tuning, applied to 3D point cloud Transformers, and achieves good performance in classification and simple part segmentation tasks.

However, existing prompt tuning methods have not fully realized the potential of standard Transformers in dense prediction tasks like 3D object detection and semantic segmentation, leading to performance bottlenecks. Standard Transformers lack inductive biases and struggle with multi-scale local information, limiting their effectiveness in these tasks. To address this, this paper proposes a novel prompt tuning framework that enhances standard Transformers’ performance in dense prediction tasks through refined feature extraction and prompt-refined learning blocks.

\section{PCT-Prompt}
\label{sec::PCT-Prompt}
\begin{figure*}[!htbp]
	\centering
	\includegraphics[width=0.9\linewidth]{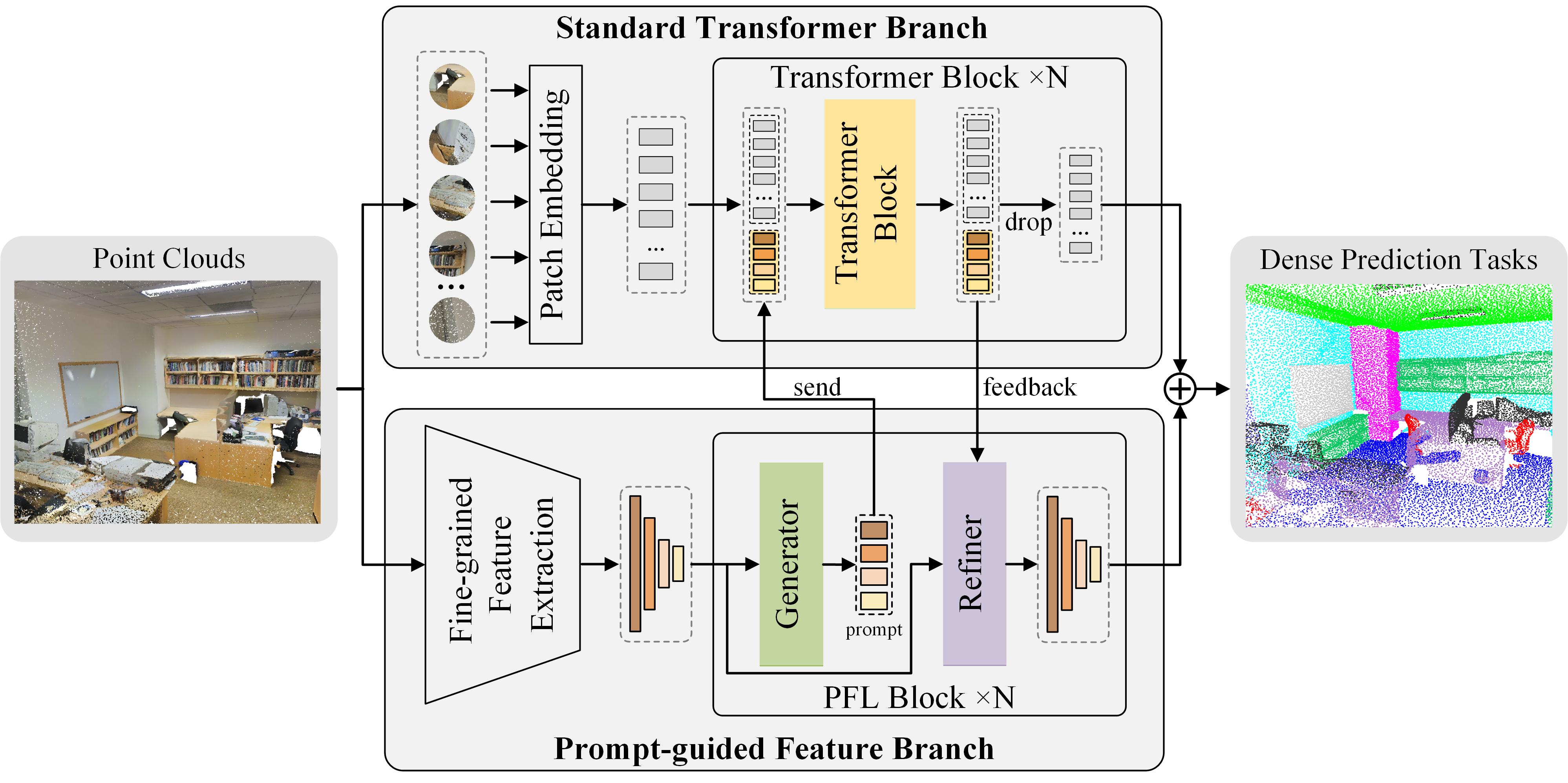}
	\caption{Overall architecture of PCT-Prompt.}
	\label{fig:method_framework}
\end{figure*}

As shown in Fig.~\ref{fig:method_framework}, the PCT-Prompt framework consists of two main branches: the standard Transformer branch and the prompt-guided feature branch. 
(1) The standard Transformer branch is capable of loading weights from various pre-trained models. It takes sub-clouds, which are obtained using Farthest Point Sampling (FPS) and K-Nearest Neighbor (KNN) algorithms, as input. By employing patch embedding, it constructs input sequences, which are then processed by a backbone comprising \(N\) Transformer blocks of the same scale. 
(2) The prompt-guided feature branch is composed of two key components: the Fine-grained Feature Extraction (FFE) block and the Prompt-refined Feature Learning (PFL) block. The FFE block employs a hierarchical structure to extract detailed geometric features from the point cloud, organizing them into a multi-scale feature pyramid. The PFL block, consisting of \(N\) blocks similar to the Transformer backbone, enables bidirectional feature interaction with the Transformer block, introducing point cloud-specific prompt features into the backbone and facilitating updates to the multi-scale feature pyramid for enhanced dense prediction tasks.

\subsection{Transformer Branch}
Although existing pre-trained Transformers, such as Point-BERT~\cite{yu2022point}, Point-MAE~\cite{pang2022masked}, ACT~\cite{dong2022autoencoders}, and ReCon~\cite{qi2023contrast}, employ different pre-training strategies, they share a common Transformer backbone and a unified pre-training parameter schema. 
To effectively utilize the weights from multiple pre-training strategies, the proposed PCT-Prompt adopts the same standard Transformer backbone for point cloud feature extraction. The following section provides a detailed overview of the standard Transformer backbone used in PCT-Prompt.

\subsubsection{Patch Embedding}
Given an input point cloud \(\mathbf{P}\in\mathbb{R}^{n\times 3}\) , the FPS algorithm is employed to select \(S\) center points, denoted as \(\widehat{\mathbf{P}}\). 
For each center point, its local neighborhood is constructed by aggregating \(K\) nearest points using the KNN algorithm, resulting in \(S\) local sub-clouds. 
To mitigate the influence of global coordinate attributes and ensure the invariance of local sub-clouds, each sub-cloud is translated by subtracting its corresponding center point. 
Subsequently, a lightweight PointNet~\cite{qi2017pointnet} is utilized to project these localized sub-clouds into point cloud embeddings. This process can be formally expressed as:
\begin{equation}
	\label{eq:mini_pointnet}
	\mathbf{F}_\text{emb} = \text{MaxPool} \left( \text{MLP} \left( \text{KNN} \left( \text{FPS} \left( \mathbf{P} \right) \right) - \widehat{\mathbf{P}} \right) \right).
\end{equation}
here, \(\text{MaxPool}(\cdot)\) denotes channel-wise max pooling; \(\text{MLP}(\cdot)\) comprises 1D convolutions (\(1\times1\) convolutions) followed by Batch Normalization and ReLU activation functions; \(\text{KNN}(\cdot)\) denotes the KNN algorithm; and \(\text{FPS}(\cdot)\) denotes for the FPS algorithm, which ensures uniform point distribution.

Given the pivotal role of positional embeddings in the Transformer architecture, we incorporate positional information into the \(S\) patch-wise features. To achieve this, we utilize a multilayer perceptron (MLP) comprising two linear layers with a GELU activation function, which transforms the \(S\) center coordinates into \(D\)-dimensional vectors, thereby generating the positional embeddings \(\mathbf{F}_\text{pos}\). Subsequently, the point cloud embeddings are combined with the positional embeddings to form the input to the standard Transformer, denoted as \(\mathbf{F}_\text{st}^{0}\), as expressed in the following equation:
\begin{equation}
	\label{eq:Patch}
	\mathbf{F}_\text{st}^{0} = \mathbf{F}_\text{pos} \oplus \mathbf{F}_\text{emb},
\end{equation}
where \(\oplus\) denotes the concatenation operator.

\subsubsection{Transformer Backbone}
The standard Transformer backbone, consisting of \(L\) uniform-scale Transformer layers, is organized into \(N\) blocks. The input to the \(i\)-th Transformer block is denoted as \(\mathbf{F}_\text{st}^{i-1} \in \mathbb{R}^{S \times D}\).

To enhance the Transformer backbone's ability to learn features beneficial for dense prediction tasks, we introduce the prompt tokens learned by the prompt-guided feature branch into the backbone. Specifically, during the \(i\)-th feature interaction between the prompt generator and the Transformer, the prompt tokens \(\mathbf{F}_\text{prompt}^i\) are combined with the feature tokens \(\mathbf{F}_\text{st}^{i-1}\) and passed together into the \(i\)-th Transformer block. The output of the \(i\)-th Transformer block is given by:
\begin{equation}
	\label{eq:Backbone}
	\left[ \widehat{\mathbf{F}}_\text{st}^i; \widehat{\mathbf{F}}_\text{prompt}^i \right] = \text{Block} \left(\left[ \mathbf{F}_\text{st}^{i-1} ; \mathbf{F}_\text{prompt}^i \right] \right),
\end{equation}
where \(\text{Block}(\cdot)\) represents the operation of the Transformer block in the backbone. Analogous to the class token in a standard Transformer, the prompt tokens not only encapsulate the global features of the point cloud but also guide the Transformer backbone by directing its focus toward local fine-grained details.

Following the configuration of the pre-trained standard Transformer, we set \(L = 12\), with the FPS sampling number \(S\) and the KNN parameter \(K\) adaptively adjusted based on the specific requirements of the dense prediction task at hand. The number of Transformer blocks, \(N\), is chosen to be 6, based on empirical observations regarding the impact of the number of interactions on model performance.

\subsection{Prompt-guided Feature Branch}
The prompt-guided feature branch improves dense prediction tasks by extracting multi-scale geometric features through the Fine-grained Feature Extraction (FFE) block and refining them with prompt tokens in the Prompt-refined Feature Learning (PFL) block. Next, we provide a detailed description of these two blocks.

\subsubsection{Fine-grained Feature Extraction}
\begin{figure}[!htbp]
	\centering
	\includegraphics[width=0.85\linewidth]{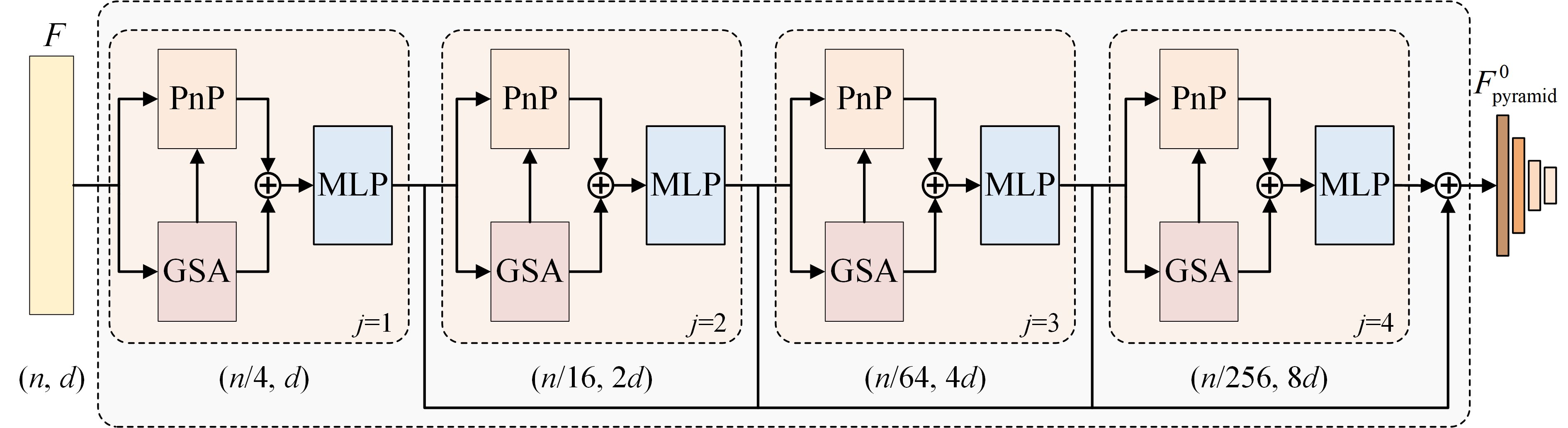}
	\caption{The structure of the fine-grained feature extraction block.}
	\label{fig:method_FFE}
\end{figure}

The proposed Fine-grained Feature Extraction (FFE) block is illustrated in Fig.~\ref{fig:method_FFE}. Initially, the input point cloud \( \mathbf{P} \in \mathbb{R}^{n \times 3} \) is encoded into feature representations \( \mathbf{F} \in \mathbb{R}^{n \times d} \) using a Multi-Layer Perceptron (MLP). A hierarchical structure with four levels is designed, where each level leverages a geometry-sensitive abstraction (GSA) layer followed by a PnP-3D~\cite{qiu2021pnp} layer. These components extract point cloud features at multiple scales, specifically \( C = \{n/4, n/16, n/64, n/256\} \). Finally, the extracted features are aggregated into a unified feature pyramid.

For the \( j \)-th feature extraction layer, FPS and KNN operations are applied to the input coordinates \( \mathbf{P}^{j-1} \in \mathbb{R}^{C^{j-1} \times 3} \) and features \( \mathbf{F}^{j-1} \in \mathbb{R}^{C^{j-1} \times D^{j-1}} \), selecting center points and their corresponding neighborhoods. Relative positional encoding~\cite{zhang2023parameter} is introduced to capture geometric relationships between the center points and neighbors. To enhance robustness, particularly in handling sparse and irregular structures, a learnable geometric affine mechanism is adopted. The output of this stage is defined as:
\begin{equation}
	\label{eq:GSA}
	\left[ \mathbf{P}^j; \mathbf{F}_\text{GSA}^j \right]= \text{GSA} \left( \left[\mathbf{P}^{j-1}; \mathbf{F}^{j-1}\right] \right),
\end{equation}
where \( \text{GSA}(\cdot) \) denotes the geometry-sensitive abstraction layer. 
The features are further refined by PnP-3D, a lightweight module that integrates local context fusion and global bilinear regularization. The local branch captures fine-grained geometric structures via neighborhood graph aggregation (e.g., EdgeConv), while the global branch enhances feature consistency through bilinear interactions across channel and point dimensions.
Finally, an MLP is applied to fuse the outputs of the GSA and PnP-3D layers, generating the point cloud features at the \( j \)-th scale:
\begin{equation}
	\label{eq:PnP3D}
	\mathbf{F}^j = \text{MLP} \left( \mathbf{F}_\text{GSA}^j \oplus \text{PnP3D} \left( \left[\mathbf{P}^j; \mathbf{F}_\text{GSA}^j \right] \right) \right),
\end{equation}
where \( \text{PnP3D}(\cdot) \) represents the PnP-3D layer, and \( \oplus \) represents the concatenation operation. 

Finally, the outputs from different scales are concatenated and transformed into a unified multi-scale feature pyramid \( \mathbf{F}_\text{pyramid}^0 \in\mathbb{R}^{\left(\frac{n}{4},\frac{n}{16},\frac{n}{64},\frac{n}{256}\right)\times D} \).

\subsubsection{Prompt-refined Feature Learning}
\begin{figure}[!htbp]
	\centering
	\includegraphics[width=0.9\linewidth]{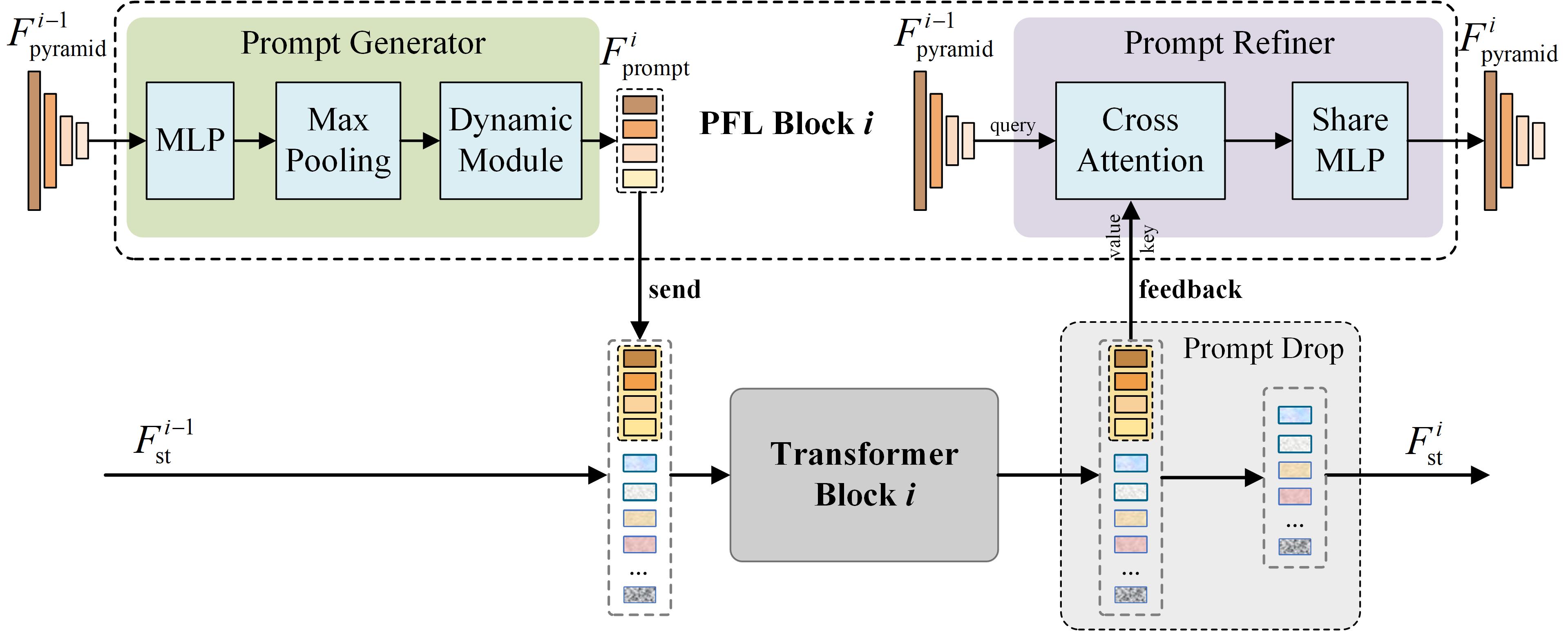}
	\caption{The structure of the prompt-refined feature learning block.}
	\label{fig:method_PFL}
\end{figure}

As shown in Fig.~\ref{fig:method_PFL}, the feature interaction between the Prompt-refined Feature Learning (PFL) block and the Transformer block is bidirectional, consisting of both sending and feedback phases. 
In the sending phase, the prompt generator injects multi-scale prompt tokens into the Transformer backbone, thereby augmenting its capacity to construct detailed feature representations. 
In the feedback phase, the Transformer block returns global feature information to the prompt refiner, enabling the feature pyramid to integrate high-level semantic knowledge and produce more refined multi-scale representations. 
A prompt drop mechanism is also introduced to progressively remove prompt tokens layer by layer, ensuring its influence extends effectively to deeper Transformer blocks and enhances feature modeling for dense segmentation tasks.

\textbf{Prompt Generator.}
The prompt generator is designed to convert the multi-scale geometric features extracted in the previous stage into task-specific prompt tokens, which are subsequently passed into the Transformer backbone. 
This process consists of two main steps:

First, adaptive max pooling is applied to extract global representations from the multi-scale feature pyramid \( \mathbf{F}_\text{pyramid}^{i-1} \), which are subsequently combined with learnable positional encodings \( \mathbf{E}_\text{pos}^i \) to yield the global feature representation \( \mathbf{F}_\text{global}^i \), as expressed in the following equation:
\begin{equation}
	\label{eq:AdaptiveMaxPool1D}
	\mathbf{F}_\text{global}^i = \text{AdaptiveMaxPool1D}\left(\text{MLP}\left(\mathbf{F}_\text{pyramid}^{i-1}\right)\right) + \mathbf{F}_\text{pos}^i.
\end{equation}
here, \(\text{MLP}(\cdot)\) comprises 1D convolutions (\(1\times1\) convolutions) followed by Batch Normalization and ReLU activation functions; \(\text{AdaptiveMaxPool1D}(\cdot)\) denotes the adaptive max pooling operation, which compresses the spatial dimensions of the multi-scale input features from \{\(\mathbb{R}^{\frac{n}{4}\times D}\), \(\mathbb{R}^{\frac{n}{16}\times D}\), \(\mathbb{R}^{\frac{n}{64}\times D}\), \(\mathbb{R}^{\frac{n}{256}\times D}\)\} to \{\(\mathbb{R}^{1\times D}\), \(\mathbb{R}^{1\times D}\), \(\mathbb{R}^{1\times D}\), \(\mathbb{R}^{1\times D}\)\}, thereby producing compact global descriptors from each scale.

Second, the global features are refined and normalized through a dynamic processing module, formulated as:
\begin{equation} 
	\label{eq:RefineAndNorm} 
	\mathbf{F}_\text{prompt}^i = \text{LayerNorm}\left(\text{LeakyReLU}\left(\text{Conv1D}\left(\mathbf{F}_\text{global}^i\right)\right)\right). 
\end{equation}
Here, \(\text{Conv1D}(\cdot)\) denotes a one-dimensional convolution (\(1\times1\) convolution) that performs linear transformation and feature reconstruction on local representations; \(\text{LeakyReLU}(\cdot)\) is a nonlinear activation function; and \(\text{LayerNorm}(\cdot)\) applies normalization across the channel dimension for each point feature.

As illustrated in Eq.~\ref{eq:Backbone}, the prompt tokens \( \mathbf{F}_\text{prompt}^i \) are integrated with the feature tokens \( \mathbf{F}_\text{st}^{i-1} \) extracted from the \( (i-1) \)-th Transformer block, serving as the input tokens to the \( i \)-th Transformer block.

\textbf{Prompt Refiner.} 
The feature interaction from the Transformer block to the prompt refiner utilizes cross-attention to refine features at each scale within the feature pyramid. Additionally, a shared-weight Multi-Layer Perceptron (MLP) is employed to efficiently propagate information across scales.

Specifically, the output feature \( [ \widehat{\mathbf{F}}_\text{st}^i; \widehat{\mathbf{F}}_\text{prompt}^i ] \) from the \( i \)-th Transformer block serves as the key and value, while the multi-scale feature pyramid \( \mathbf{F}_\text{pyramid}^{i-1} \) acts as the query. The integration of global features into the multi-scale feature pyramid through cross-attention is formulated as:
\begin{equation} 
	\label{eq:MCA_softmax}
	\widehat{\mathbf{F}}_\text{pyramid}^i = \text{Softmax}\bigg( \frac{\mathrm{LN}\left( \mathbf{F}_\text{pyramid}^{i-1} \right) \cdot \mathrm{LN}\left( \left[ \widehat{\mathbf{F}}_\text{st}^i; \widehat{\mathbf{F}}_\text{prompt}^i \right] \right)^T}{\sqrt{d}} \bigg) \cdot \mathrm{LN}\left( \left[ \widehat{\mathbf{F}}_\text{st}^i; \widehat{\mathbf{F}}_\text{prompt}^i \right] \right).
\end{equation}

Here, \(\text{Softmax}(\cdot)\) normalizes the attention scores into a probability distribution, \( \mathrm{LN}(\cdot) \) denotes layer normalization, and \( d \) represents the dimensionality of the query and key vectors, which corresponds to the feature dimension \( D \) introduced previously.

The shared-weight MLP further refines the propagation of features across scales, as described by:
\begin{equation}
	\label{eq:MLPshared}
	\mathbf{F}_\text{pyramid}^i = \mathrm{MLP}_\text{shared}\left(\widehat{\mathbf{F}}_\text{pyramid}^i\right),
\end{equation}
where \( \mathrm{MLP}_\text{shared}(\cdot) \) represents the shared-weight multi-layer perceptron. The reorganized multi-scale feature pyramid results in a more robust representation, thereby enhancing the Transformer's ability to capture detailed point cloud features and adapt effectively to dense prediction tasks.

\textbf{Prompt Drop.} 
As the number of Transformer blocks increases, each subsequent block's ability to effectively utilize prompt features diminishes, which restricts feature modeling in dense segmentation tasks. To address this, we introduce the prompt drop mechanism, ensuring that each Transformer block leverages prompt information effectively while preserving global semantic consistency and refining local geometric details.

As shown in Fig.~\ref{fig:method_PFL}, the output features of the \( i \)-th Transformer block consist of two components: prompt tokens \( \widehat{\mathbf{F}}_\text{prompt}^i \) and feature tokens \( \widehat{\mathbf{F}}_\text{st}^i \). The prompt drop mechanism removes the prompt tokens, with the final output features of the \( i \)-th Transformer block defined as:
\begin{equation}
	\label{eq:PromptDrop}
	\mathbf{F}_\text{st}^i = [ \widehat{\mathbf{F}}_\text{st}^i; \widehat{\mathbf{F}}_\text{prompt}^i ] \setminus \widehat{\mathbf{F}}_\text{prompt}^i,
\end{equation}
where \( \setminus \) represents the removal operation. By progressively eliminating prompt tokens, the mechanism ensures their influence is propagated to deeper Transformer blocks, thereby enhancing the network’s feature modeling capability for dense segmentation.

\subsection{Feature Fusion and Dense Prediction.}
The fusion process between the output features of the standard Transformer branch and the prompt-guided feature branch is designed to integrate global semantic context with local geometric details. Specifically, the output features from the Transformer branch are first upsampled to four different resolutions using the point feature upsampling strategy, generating a set of multi-scale features that are resolution-aligned with the output feature from the prompt-guided feature branch. The corresponding multi-scale features from the standard Transformer branch and the prompt-guided feature branch are then combined via element-wise addition to obtain the fused representation. This fused representation is subsequently passed to a decoder module, which is structurally analogous to the segmentation head in PointNet++~\cite{qi2017pointnet++}. The decoder comprises a series of feature propagation layers that integrate interpolation and MLPs to progressively recover fine-grained point-level semantic predictions.

\subsection{Loss Function}
We adopt two loss functions tailored to the characteristics of the datasets: cross-entropy loss and weighted cross-entropy loss. For the ShapeNetPart dataset, which features a balanced class distribution, we employ standard cross-entropy loss to quantify the divergence between the predicted and ground truth class distributions. It is defined as:
\begin{equation}
	\label{eq::loss_ce}
	\mathcal{L}_{\text{ce}} = - \sum_{i=1}^{C} y_i \log(\hat{y}_i),
\end{equation}
where \( y_i \) denotes the ground truth label for class \(i\), and \( \hat{y}_i \) represents the predicted probability for class \(i\). 
For the S3DIS and DALES datasets, which are characterized by significant class imbalance, we utilize weighted cross-entropy loss to address this challenge. The loss function is expressed as:
\begin{equation}
	\label{eq::loss_wce}
	\mathcal{L}_{\text{wce}} = - \sum_{i=1}^{C} w_i \cdot y_i \log(\hat{y}_i),
\end{equation}
where \( w_i \) is the weight assigned to class \(i\), calculated as the ratio of the number of points in class \(i\) (\( f_i \)) to the total number of points across all classes (\( N \)), i.e., \( w_i = \frac{N}{f_i} \). This weighting mechanism assigns higher importance to less frequent classes, thereby mitigating the bias towards dominant classes and enhancing the model's performance on imbalanced datasets.

\section{Experiments}
\label{sec::Experiments}
\subsection{Experiments Setting}
\subsubsection{ShapeNetPart}
The ShapeNetPart dataset~\cite{chang2015shapenet} contains 16,881 objects, spanning 16 categories and 50 part labels, with each instance consisting of 2 to 6 parts. The model setup strictly follows that of Point-BERT. For each object, 2,048 points are randomly sampled as input. The standard Transformer backbone is configured with a sampling count \(S=128\), corresponding to 128 point cloud patches, and the KNN parameter \(K\) is set to 32. The Prompt reduces the number of GSA layers to 3, with the downsampling ratio set to \(\mathbf{C}=\{{n\over8},{n\over16},{n\over32}\}\), while maintaining the same KNN parameters as the backbone.

For the training setup, we use cross-entropy loss (Eq.~\ref{eq::loss_ce}) to measure the discrepancy between the predicted probability distribution for each point and the ground truth labels. The model is trained for 300 epochs on an NVIDIA RTX 4090 with 24GB of memory using the AdamW optimizer, with a learning rate of 0.0005 and a batch size of 6. The quantitative evaluation metrics include class mIoU, instance mIoU, and the IoU for each individual class.

\subsubsection{S3DIS}
We validate the semantic segmentation performance of PCT-Prompt on the real-world indoor scene S3DIS dataset~\cite{armeni20163d}. S3DIS includes six large indoor areas from three different buildings, with a total of 273 million points annotated with 13 categories (ceiling, floor, table, etc.). Area 5, the most challenging region, is used for testing, while the remaining areas are used for training.

For model configuration, due to the large point cloud scene data, we randomly sample 12,000 points as the input for each scene. The standard Transformer backbone is set with a sampling number \(S=256\), corresponding to 256 point cloud patches, and the KNN parameter \(K=32\). In the Prompt, the number of GSA layers is set to 4, and the downsampling ratios are \(\mathbf{C}=\{{n\over4},{n\over16},{n\over64},{n\over256}\}\), with KNN set to 64.

For training, due to the significant class imbalance in the real-world dataset, we use a weighted cross-entropy loss (Eq.~\ref{eq::loss_wce}) to address the class imbalance issue. The class weights are calculated based on the frequency of each class in the input point clouds. The model is trained for 150 epochs using the AdamW optimizer with a learning rate of 0.0005 and a batch size of 6 on an NVIDIA RTX 4090 with 24GB of memory. The quantitative evaluation metrics include mIoU, mAcc, OA, and IoU for each class.

\subsubsection{DALES}
DALES is a 10 km$^2$ airborne LiDAR dataset that includes 40 urban and rural scenes with a total of 500 million points, of which 12 scenes are used for evaluation.

For model configuration, due to the large number of points in each frame, each scene is divided into multiple grids of size \(40\times40\) based on the area. 20,000 points from each point cloud scene are selected as input. The standard Transformer backbone is set with a sampling number \(S=384\), corresponding to 384 point cloud patches, and the KNN parameter \(K=32\). In the Prompt, the number of GSA layers is set to 4, and the downsampling ratios are \(\mathbf{C}=\{{n\over4},{n\over16},{n\over64},{n\over256}\}\), with KNN set to 64.

For training, weighted cross-entropy loss (Eq.~\ref{eq::loss_wce}) is used to address class imbalance, where the class weights are derived from the frequency of each class in the labels. The model is trained for 250 epochs using the AdamW optimizer with a learning rate of 0.0005 and a batch size of 1 on an NVIDIA RTX 4090 with 24GB of memory. The quantitative evaluation metrics include mIoU and IoU for each class.

\subsection{Quantitative Analysis}
\subsubsection{Performance on ShapeNetPart}
We leveraged pre-trained weights from Point-BERT~\cite{yu2022point} for both the standard Transformer (Point-BERT) and PCT-Prompt, conducting part segmentation on the ShapeNetPart dataset. The experimental results, presented in Table~\ref{tab:part_segmentation}, reveal that, when using identical pre-trained parameters for the standard Transformer backbone, PCT-Prompt enhances part segmentation performance. Specifically, it improves the category mIoU and instance mIoU by $0.9\%$ and $0.6\%$, respectively, by incorporating the prompt structure. Moreover, PCT-Prompt effectively reduces the performance gap between the standard Transformer and its variants, such as PCT~\cite{guo2021pct} and pointCAT~\cite{yang2023pointcat}. These findings demonstrate that PCT-Prompt introduces an effective task-specific prompting mechanism, substantially benefiting the standard Transformer architecture.
\begin{table*}[!htbp]
	\centering
	\caption{Object part segmentation results on the ShapeNetPart dataset. “T”, “S”, and “V” represent traditional neural network models, standard Transformers, and variant Transformers, respectively.}
	\label{tab:part_segmentation}
	\resizebox{\textwidth}{!}{
		\begin{tabular}{l|c|cc|cccccccccccccccc}
			\toprule
			Methods &Type	&\makecell{ins.\\mIoU} &\makecell{cls.\\mIoU} &aero &bag 	&cap 	&car 	&chr. 	&e.ph. 	&gtar 	&knf. 	&lmp. 	&ltp. 	&m.b. 	&mg 	&pstl 	&s.b. 	&skte 	&tbl.\\
			\midrule
			PointNet\cite{qi2017pointnet}		&T	&83.7&80.4	&83.4	&78.7	&82.5	&74.9	&89.6	&73.0	&91.5	&85.9	&80.8	&95.3	&65.2	&93.0	&81.2	&57.9	&72.8	&80.6\\
			PointNet++\cite{qi2017pointnet++}	&T	&85.1&81.9	&82.4	&79.0	&87.7	&77.3	&90.8	&71.8	&91.0	&85.9	&83.7	&95.3	&71.6	&94.1	&81.3	&58.7	&76.4	&82.6\\
			PointCNN\cite{li2018pointcnn}		&T	&86.1&84.6	&84.1	&86.5	&86.0	&80.8	&90.6	&79.7	&92.3	&88.4	&85.3	&96.1	&77.2	&95.2	&84.2	&64.2	&80.0	&83.0\\
			DGCNN\cite{wang2019dynamic}			&T	&85.2&82.3	&84.0	&83.4	&86.7	&77.8	&90.6	&74.7	&91.2	&87.5	&82.8	&95.7	&66.3	&94.9	&81.1	&63.5	&74.5	&82.6\\
			PointMLP\cite{ma2022rethinking}		&T	&86.1&84.6	&83.5	&83.4	&87.5	&80.5	&90.3	&78.2	&92.2	&88.1	&82.6	&96.2	&77.5	&95.8	&85.4	&64.6	&83.3	&84.3\\
			PointASNL\cite{yan2020pointasnl}	&T	&86.1&-	&84.1	&84.7	&87.9	&79.7	&92.2	&73.7	&91.0	&87.2	&84.2	&95.8	&74.4	&95.2	&81.0	&63.0	&76.3	&83.2\\
			PointCAT\cite{yang2023pointcat}		&V	&86.0&84.4	&83.0	&83.8	&90.1	&79.8	&90.2	&83.4	&91.8	&87.8	&82.5	&95.9	&76.1	&95.4	&84.9	&68.5	&83.1	&84.1\\
			PCT\cite{guo2021pct}				&V	&86.4&-	&85.0	&89.0	&82.4	&81.2	&91.9	 &71.5	&91.3	 &88.1	&86.3	&95.8	&64.6	&95.8	&83.6	&62.2	&77.6	&83.7\\
			PT\cite{zhao2021point}				&V	&86.6&83.7	&-	&-	&-	&-	&-	&-	&-	&-	&-	&-	&-	&-	&-	&-	&-	&-\\
			\midrule
			Point-BERT\cite{yu2022point}		&S	&85.6&84.1	&84.3	&84.8	&88.0	&79.8	&91.0	&81.7	&91.6	&87.9	&85.2	&95.6	&75.6	&94.7	&84.3	&63.4	&76.3	&81.5\\
			PCT-Prompt							&S &86.2&85.0	&83.8	&85.1	&90.5	&80.0	&90.5	&83.6	&92.1	&88.0	&83.5	&96.0	&77.1	&94.7	&85.6	&64.4	&81.0	&83.9\\
			\bottomrule
	\end{tabular}}
\end{table*}

\subsubsection{Performance on S3DIS}
We initialize both the standard Transformer and PCT-Prompt with the same pre-trained parameters and evaluate their performance on the real-world indoor dataset S3DIS for semantic segmentation. As shown in Table~\ref{tab:segmentation_S3DIS}, when both models are initialized with Point-BERT pre-trained weights, PCT-Prompt outperforms the standard Transformer, achieving a $3.1\%$ improvement in mAcc and a $7.4\%$ improvement in mIoU. Furthermore, PCT-Prompt surpasses several state-of-the-art point cloud Transformer variants, including Superpoint Transformer~\cite{sun2023superpoint}, PCT~\cite{guo2021pct}, pointCAT~\cite{yang2023pointcat}, PatchFormer~\cite{zhang2022patchformer}, and pointTransformer~\cite{zhao2021point}. These results not only validate the design principles underlying PCT-Prompt but also highlight the effectiveness of the prompt mechanism in extending the standard Transformer to a diverse range of downstream tasks, thereby enabling more effective utilization of the feature representation capabilities of the general backbone.

\begin{table*}[!htbp]
	\centering
	\caption{Semantic segmentation results on S3DIS dataset, tested on Area 5. *means we report the best results among public codebases and our own reproduction.}
	\label{tab:segmentation_S3DIS}
	\resizebox{\textwidth}{!}{
		\begin{tabular}{l|c|ccc|ccccccccccccc}
			\toprule
			Methods 								&Type	&mIoU 	&mAcc 	&OA 	&ceil. 	&flr 	&wl 	&bm 	&col. 	&wnd. 	&dr 	&tbl. 	&chr 	&sf. 	&b.c. 	&brd 	&cltr.\\
			\midrule
			PointNet\cite{qi2017pointnet}			&T&41.1 	&49.0 	&-	&88.8 	&97.3 	&69.8 	&0.1 	&3.9 	&46.3 	&10.8 	&59.0 	&52.6 	&5.9 	&40.3 	&26.4 	&33.2\\
			%SegCloud\cite{tchapmi2017segcloud}		&T&48.9 	&57.4 	&-	&90.1 	&96.1 	&69.9 	&0.0 	&18.4 	&38.4 	&23.1 	&70.4 	&75.9 	&40.9 	&58.4 	&13.0 	&41.6\\
			%TangentConv\cite{tatarchenko2018tangent}&T&52.6 	&62.2 	&-	&90.5 	&97.7 	&74.0 	&0.0 	&20.7 	&39.0 	&31.3 	&77.5 	&69.4 	&57.3 	&38.5 	&48.8 	&39.8\\
			PointCNN\cite{li2018pointcnn}			&T&57.3 	&63.9 	&85.9 	&92.3 	&98.2 	&79.4 	&0.0 	&17.6 	&22.8 	&62.1 	&74.4 	&80.6 	&31.7 	&66.7 	&62.1 	&56.7\\
			SPG\cite{landrieu2018large}				&T&58.0 	&66.5 	&86.4 	&89.4 	&96.9 	&78.1 	&0.0	&42.8 	&48.9 	&61.6 	&84.7 	&75.4 	&69.8 	&52.6 	&2.1	&52.2\\		
			%PCCN\cite{wang2018deep}					&T&58.3 	&67.0 	&-	&92.3 	&96.2 	&75.9 	&0.3 	&6.0 	&69.5 	&63.5 	&66.9 	&65.6 	&47.3 	&68.9 	&59.1 	&46.2\\
			PointWeb\cite{zhao2019pointweb}			&T&60.3 	&66.6 	&87.0 	&92.0 	&98.5 	&79.4 	&0.0	&21.1 	&59.7 	&34.8 	&76.3 	&88.3 	&46.9 	&69.3 	&64.9 	&52.5\\
			KPConv\cite{thomas2019kpconv}			&T&67.1 	&72.8 	&-	&92.8 	&97.3 	&82.4 	&0.0 	&23.9 	&58.0 	&69.0 	&81.5 	&91.0 	&75.4 	&75.3 	&66.7 	&58.9\\
			%DSP\cite{wei2023dense}					&T&68.5	&-		&-	&94.5	&98.4	&83.7	&0.0	&28.9	&59.1	&73.4	&91.2	&82.2	&75.3	&73.9	&68.5	&61.0\\
			PAT\cite{yang2019modeling}				&V&60.1 	&70.8 	&-	&93.0 	&98.5 	&72.3 	&1.0 	&41.5 	&85.1 	&38.2 	&57.7 	&83.6 	&48.1 	&67.0 	&61.3 	&33.6\\
			SPT\cite{sun2023superpoint}				&V&68.9	&77.3 	&89.5 	&91.5 	&98.2 	&81.4 	&0.0	&23.3 	&65.3 	&40.0 	&75.5 	&87.7 	&58.5 	&67.8 	&65.6 	&49.4\\
			PatchFormer\cite{zhang2022patchformer}	&V&67.3	&-		&- 	&91.8 	&98.7 	&86.2 	&0.0	&34.1 	&48.9 	&62.4 	&81.6 	&89.8	&47.2 	&74.9 	&74.4 	&58.6\\
			PT*\cite{zhao2021point}					&V&70.0	&76.8 	&90.4 	&94.0 	&98.5 	&86.3 	&0.0 	&38.0 	&63.4 	&74.3 	&89.1 	&82.4 	&74.3 	&80.2 	&76.0 	&59.3\\
			PCT\cite{guo2021pct}					&V&61.3 	&67.7 	&-	&92.5	&98.4	&80.6	&0.0	&19.3	&61.6	&48.0	&76.6	&85.2	&46.2	&67.7	&67.9	&52.3\\
			PointCAT\cite{yang2023pointcat}			&V&64.0	&71.0 	&88.2	&94.2	&98.3	&80.5	&0.0	&18.6	&55.5	&58.9	&77.2	&88.0	&64.8	&72.2	&68.9	&55.4\\
			\midrule
			Point-BERT\cite{yu2022point}			&S&63.5 	&75.7 	&85.0 	&91.3 	&92.3 	&73.1 	&0.0	&33.9 	&65.6 	&60.4 	&76.5 	&82.7 	&86.8 	&64.0 	&41.7 	&43.0\\
			PCT-Prompt								&S&70.9	&78.8	&90.3	&95.2	&98.1	&83.6	&0.0	&63.4	&82.2	&71.1	&78.5	&86.8	&76.8	&70.8	&59.2	&55.4\\
			\bottomrule
	\end{tabular}}
\end{table*}

\begin{figure*}[!htbp]
	\centering
	\includegraphics[width=1\linewidth]{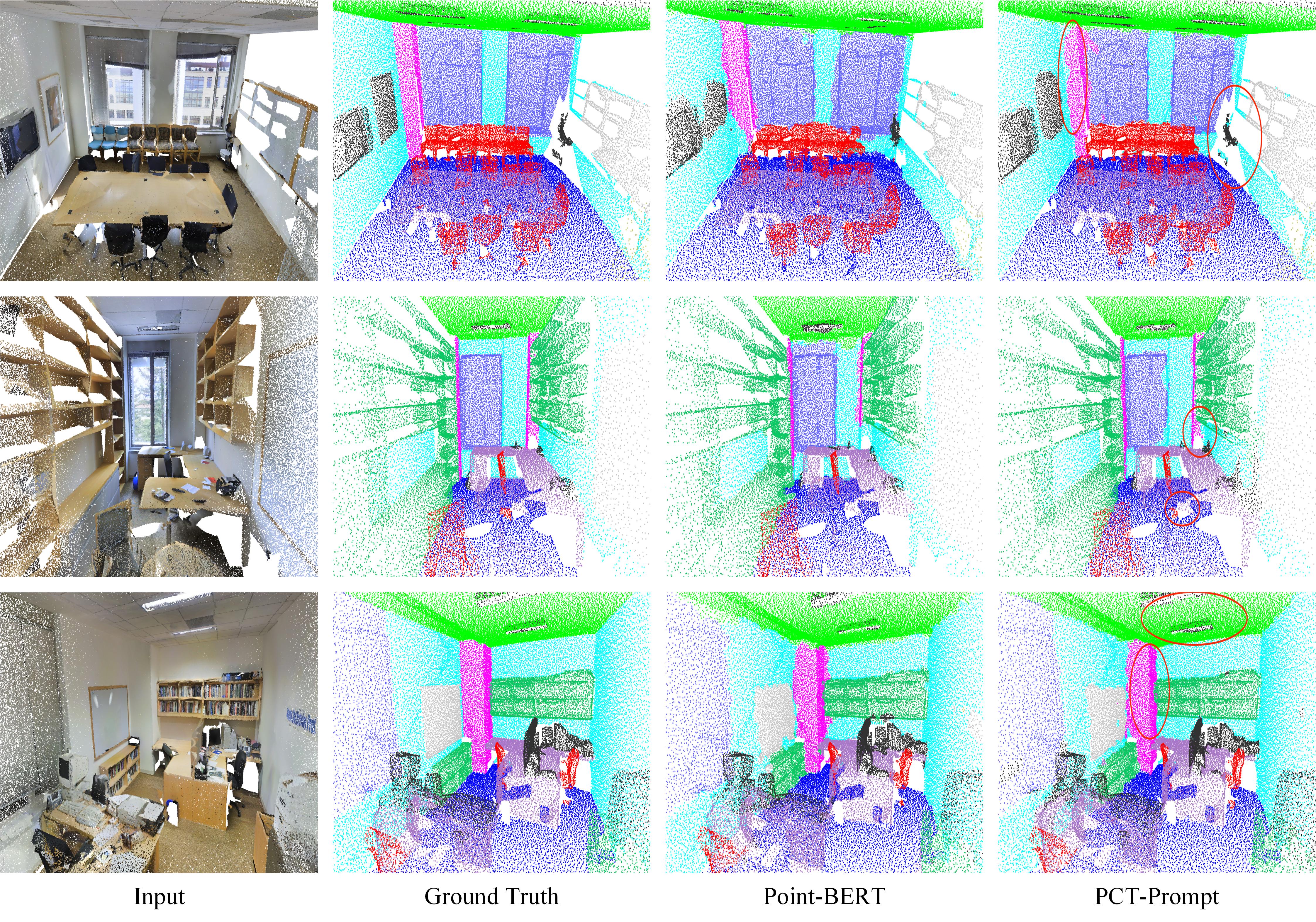}
	\caption{Visualizing the comparison between Point-BERT and PCT-Prompt on the S3DIS Dataset.}
	\label{fig:visual_S3DIS}
\end{figure*}

Furthermore, we visually compare the results of the standard Transformer (Point-BERT) and PCT-Prompt across three scenes in Area 5. 
As depicted in Fig.~\ref{fig:visual_S3DIS}, PCT-Prompt demonstrates a clear fine-grained enhancement effect on the Point-BERT. 
These results show that the PCT-Prompt effectively extends the standard Transformer to downstream tasks, proves the rationality of the PCT-Prompt design, and can deeply explore the feature extraction potential of the general point cloud Transformer backbone.

\subsubsection{Performance on DALES}
We evaluated both the standard Transformer and PCT-Prompt models using the same pre-trained parameters on the real-world DALES dataset for semantic segmentation. 
As shown in Table~\ref{tab:dales_results}, PCT-Prompt outperforms the standard Transformer by achieving a $6.5\%$ improvement in mIoU. This performance boost in more complex and diverse real-world scenarios can be attributed to PCT-Prompt's effective integration of multi-scale geometric features and its ability to complement the standard Transformer architecture, thereby enhancing the overall semantic segmentation performance.
\begin{table*}[htbp]
	\centering
	\caption{Comparison of semantic segmentation performance on the DALES dataset.}
	\label{tab:dales_results}
	\resizebox{\textwidth}{!}{
		\begin{tabular}{l|c|ccc|cccccccc}
			\toprule
			Methods &Type	&mIoU &mAcc &OA &Ground &Buildings &Cars &Trucks &Poles &Power Lines &Fences &Vegetation\\ 
			\midrule
			PointNet++\cite{qi2017pointnet++} 	&T&68.3	&- 		&95.7 	&94.1 	&89.1 	&75.4 	&30.3 	&40.0 	&79.9 	&46.2 	&91.2\\
			%ConvPoint\cite{boulch2020convpoint} 	&T&67.4	&- 		&97.2 	&96.9 	&96.3 	&75.5 	&21.7 	&40.3 	&86.7 	&29.6 	&91.9\\
			SPG\cite{landrieu2018large}			&T&60.6 	&- 		&95.5 	&94.7 	&93.4 	&62.9 	&18.7 	&28.5 	&65.2 	&33.6 	&87.9\\
			PointCNN\cite{li2018pointcnn} 	&T&58.4 	&- 		&97.2 	&97.5 	&95.7 	&40.6 	&4.8	&57.6 	&26.7 	&52.6 	&91.7\\
			%ShellNet\cite{zhang2019shellnet} 	&T&57.4 	&- 		&96.4 	&96.0 	&95.4 	&32.2 	&39.6 	&20.0 	&27.4 	&60.0 	&88.4\\ 
			%SuperCluster\cite{robert2024scalable}	&T&77.3	&- 		&-	&-&-&-&-&-&-&-&-\\
			SPT\cite{sun2023superpoint} 			&V&79.6	&- 		&97.5	&96.7	&93.1	&86.1	&52.4	&94.0	&52.7	&65.3	&96.7\\
			\midrule
			Point-BERT\cite{yu2022point} 			&S&72.0	&79.6 	&96.0	&92.9	&91.9	&63.3	&27.8	&93.3	&49.0	&65.4	&93.1\\
			PCT-Prompt								&S&78.5	&84.9	&97.3	&98.8	&96.8	&91.8	&45.9	&97.5	&68.9	&76.6	&96.9\\
			\bottomrule
	\end{tabular}}%
\end{table*}

\subsection{Ablation Studies}
\subsubsection{Components of PCT-Prompt}
The purpose of the ablation study on the PCT-Prompt component is to assess the individual contributions of each block. As shown in Table~\ref{tab:Component_ablation}, PCT-Prompt consists of three primary components: the Transformer backbone, the FFE block, and the PFL block. The PFL block is further divided into the generator, refiner, and drop mechanisms. Using only the Transformer yields a baseline mIoU of 63.5, while using the FFE block alone results in a significantly lower mIoU of 55.6, indicating that FFE block requires the Transformer backbone for effective feature modeling. 
When FFE block is combined with the Transformer backbone using the simplest interaction scheme (i.e., direct feature addition), the mIoU improves by 2.0\%, suggesting its complementary role. Adding either the generator or the refiner leads to additional gains, with the refiner providing slightly better performance. Combining both results in further improvement, and incorporating the drop mechanism yields the best overall performance, confirming the effectiveness of the complete design. These results demonstrate that each block contributes meaningfully, and their integration enhances performance in dense point cloud segmentation tasks.

\begin{table}[!htbp]
	\centering
	\caption{Component ablation study of PCT-Prompt on the S3DIS.}
	\label{tab:Component_ablation}
	%\resizebox{\textwidth}{!}{
		\begin{tabular}{ccccccc}
			\toprule
			\multirow{2}{*}{Transformer}&\multirow{2}{*}{FFE}&\multicolumn{3}{c}{PFL} &\multirow{2}{*}{mIoU}&\multirow{2}{*}{mAcc}\\
			\cmidrule (r){3-5}
			&&Generator	&Refiner	&Drop&&\\
			\midrule
			$\surd$ &	$\times$ &	$\times$ &$\times$ 	&$\times$ 	&63.5	&75.7\\
			$\times$ &	$\surd$ &	$\times$ &$\times$ 	&$\times$ 	&55.6	&65.7\\
			$\surd$ &	$\surd$ &	$\times$ &$\times$ 	&$\times$ 	&65.5	&75.8\\
			$\surd$ &	$\surd$ &	$\surd$	 &$\times$	&$\times$ 	&67.2	&76.0\\
			$\surd$	&$\surd$	 &$\times$	 &$\surd$	&$\times$ 	&66.7	&76.1\\
			$\surd$ &	$\surd$ &	$\surd$	 &$\surd$	&$\times$ 	&69.0	&77.5\\
			$\surd$ &	$\surd$ &	$\surd$	 &$\times$	&$\surd$ 	&68.3	&77.0\\
			$\surd$ &	$\surd$ &	$\surd$	 &$\surd$	&$\surd$	&\textbf{70.9}	&\textbf{78.8}\\
			\bottomrule
		\end{tabular}
\end{table}

\subsubsection{Effects of prompt Frequency}
\label{subsubsection:efibn}
To refine the feature representation capabilities, PCT-Prompt divides the Transformer backbone into \(N\) blocks and constructs \(N\) corresponding Prompt-refined Feature Learning (PFL) blocks. This multi-block structure facilitates feature interactions, improving the model’s ability to capture complex features. To evaluate the impact of interaction frequency \(N\), we conducted experiments with \(N\) values set to 0, 1, 2, 4, 6, and 8, keeping training configurations constant to determine the optimal \(N\).

Table~\ref{tab:Quantitative_interactions} shows the semantic segmentation performance on the S3DIS dataset for different interaction frequencies. When \(N = 0\), PCT-Prompt is equivalent to the standard Transformer, achieving an mIoU of $63.5\%$. As the interaction frequency increases, performance improves, peaking at \(N = 6\), where the mIoU reaches $70.9\%$. This demonstrates that increasing the frequency of interactions improves the fusion quality between the PFL block and the Transformer, with \(N = 6\) being the optimal value. These results highlight the importance of frequent feature interactions in improving PCT-Prompt’s performance.
\begin{table}[!htbp]
	\centering
	\caption{Quantitative comparisons of different numbers of interactions on the S3DIS.}
	\label{tab:Quantitative_interactions}
	\begin{tabular}{c|cc}
		\toprule
		Frequency &mIoU	&mAcc\\
		\midrule
		\(N = 0\)	&63.5&75.7	\\
		\(N = 1\)	&65.9&76.1	\\
		\(N = 2\)	&68.4&76.0	\\
		\(N = 4\)	&68.9&76.2	\\
		\(N = 6\)	&\textbf{70.9}&\textbf{78.8}	\\
		\(N = 8\)	&67.4&75.5	\\
		\bottomrule
	\end{tabular}
\end{table}
	
\subsubsection{Effects of Different Pre-training Models}
To assess the generalization capability of PCT-Prompt across various pre-trained models in dense prediction tasks, we evaluated its performance with different pre-trained weights. The pre-trained models considered in this study include ACT~\cite{dong2022autoencoders}, Point-MAE~\cite{pang2022masked}, MaskPoint~\cite{liu2022masked}, Point-BERT~\cite{yu2022point}, and ReCon~\cite{qi2023contrast}.
	
\begin{table}[!htbp]
	\centering
	\caption{Quantitative comparisons among various pre-trained models.}
	\label{tab:Quantitative_pretrained}
	\begin{tabular}{c|l|cc}
		\toprule
		Methods	 &Pre-trained Models		&mIoU	&mAcc\\
		\midrule
		\multirow{6}{*}{\makecell{standard\\Transformer}} 	&-	&72.1	&62.9\\
		&ACT\cite{dong2022autoencoders}		&63.1&73.2	\\
		&Point-MAE\cite{pang2022masked}	 	&63.1&72.1	\\
		&MaskPoint\cite{liu2022masked} 		&64.4&72.7	\\
		&Point-BERT\cite{yu2022point}		&63.5&75.7	\\
		&ReCon\cite{qi2023contrast}	 		&64.8&73.3	\\
		\midrule
		\multirow{6}{*}{PCT-Prompt}	&-		&64.2&73.7	\\
		&ACT\cite{dong2022autoencoders}	 	&65.2&75.4	\\
		&Point-MAE\cite{pang2022masked}	 	&65.1&73.9	\\
		&MaskPoint\cite{liu2022masked} 		&67.1&73.5	\\
		&Point-BERT\cite{yu2022point}		&\textbf{70.9}&\textbf{78.8}	\\
		&ReCon\cite{qi2023contrast}	 		&66.1&77.6	\\
		\bottomrule
	\end{tabular}
\end{table}
	
As shown in Table~\ref{tab:Quantitative_pretrained}, the second and eighth rows represent the performance of the standard Transformer and PCT-Prompt without pre-trained weights, respectively. The third to seventh rows correspond to the standard Transformer with various pre-trained weights, while the ninth to final rows show the performance of PCT-Prompt under the same pre-trained weight configurations. 
The comparative analysis reveals the following key insights: 
(1) The integration of prompt-based features consistently improves the performance of the standard Transformer. 
(2) All pre-trained models contribute positively to the performance of PCT-Prompt across dense prediction tasks. 
(3) Among the pre-trained models tested, Point-BERT leads to the best performance when applied to PCT-Prompt.
	
These results highlight the robustness and generalizability of PCT-Prompt in leveraging diverse pre-trained models to enhance the adaptability and performance of the standard Transformer across various dense prediction tasks. Furthermore, our experimental findings conclusively demonstrate that the proposed PCT-Prompt framework not only boosts existing standard Transformer architectures for dense prediction but also holds the potential to further amplify performance as more advanced standard Transformers are developed.

\section{Conclusion}
\label{sec::Conclusion}
This paper presents the Point Cloud Transformer Prompt (PCT-Prompt), a novel framework that enhances the capability of standard Transformers for dense prediction tasks in point cloud data. PCT-Prompt augments the feature representation of the standard Transformer without altering its core architecture, by integrating a pre-training-free feature extraction module and a prompt-guided feature learning mechanism. This framework improves model performance in dense prediction tasks by effectively incorporating multi-scale geometric features and dynamically refined prompt tokens, which interact with the Transformer backbone to iteratively enhance the feature pyramid. Moreover, PCT-Prompt is designed as a flexible and generalizable framework that can seamlessly load weights from various pre-trained models, thereby offering enhanced adaptability across diverse datasets. Through extensive experiments on segmentation tasks, we demonstrate that PCT-Prompt effectively reduces the performance gap between standard and variant Transformers, expanding the task applicability of standard Transformers and improving their effectiveness in real-world scenarios.

\bibliographystyle{unsrt}  
\bibliography{Reference}  %%% Remove comment to use the external .bib file (using bibtex).

%%%%%%%%% BODY TEXT

\end{document}